\documentclass[table]{gtech}

\usepackage{amssymb}
\usepackage{bigdelim}
\usepackage{longtable}
\usepackage{tabularray}
\usepackage{wrapfig}
\usepackage{float}
\usepackage{datatool}
\usepackage{enumitem}
\ifPDFTeX
  \usepackage{tgpagella}
  \usepackage{mathpazo}
\else
  \usepackage{unicode-math}
\fi
\usepackage{inconsolata}
\usepackage{makecell}
\usepackage{multirow}
\usepackage{adjustbox}
\usepackage{tablefootnote}
\usepackage{array}
\usepackage{nicefrac}
\usepackage{xspace}
\usepackage{amsthm}
\usepackage{amsmath}
\usepackage{arydshln}
\usepackage{pifont}
\usepackage{tabulary}
\usepackage{fontawesome5}
\usepackage{bbding}
\usepackage{multicol}

\title{Hyperparameter Scaling Laws Across MoE Sparsity}

\author[]{\vspace{0.5em}Changxin Tian}
\author[]{Kunlong Chen}
\author[]{Jia Liu}
\author[]{Ziqi Liu}
\author[\dag]{Zhiqiang Zhang}
\author[\dag]{Jun Zhou}

\affiliation[]{Ling Team, Ant Group}
\contribution[\dag]{Corresponding author}

\abstract{Mixture-of-Experts (MoE) models expand model capacity without a proportional increase in training compute, but increasing sparsity makes reliable hyperparameter transfer challenging. In this work, we show that conventional hyperparameter scaling laws are insufficient for ultra-sparse MoEs: the optimal learning rate and batch size vary with activation ratio, and these shifts cannot be explained by either total or activated parameter count alone. To characterize this dependence, we conduct {1,800 pre-training runs} spanning six activated-parameter scales and models with up to 6B total non-embedding parameters, processing approximately 20 trillion tokens at a cost of {200,000 equivalent H800 GPU-hours}. 
Our results reconcile conflicting findings in prior work by revealing two scaling regimes. At fixed sparsity, the optimal batch size follows a power-law relationship with training tokens $D$, whereas the optimal learning rate scales with training compute $C$ and remains robust to the allocation between model size and data. Across sparsity levels, the activation ratio $A$ enters both relationships as an additional multiplicative power-law factor. These observations lead to unified hyperparameter scaling laws that transfer across MoE sparsity levels. 
Large-scale evaluation shows that the scaling form outperforms alternative functional forms. On a held-out ultra-sparse MoE with 12B total parameters and only 1/64 of its experts activated, the predicted hyperparameters remain close to the observed optima, supporting joint extrapolation across model scale and sparsity. Further experiments demonstrate transfer across expert granularities and isolate the effect of activation ratio from that of total expert count.
}
\date{\today}
\gtechdata[Correspondence]{\email{\{tianchangxin.tcx,lingyao.zzq,jun.zhoujun\}@antgroup.com}}

\begin{document}
\maketitle

\begin{figure*}[!b]
    \centering
    \captionsetup{skip=5pt}
    \captionsetup[subfigure]{skip=1pt}
    \begin{subfigure}[b]{0.34\textwidth}
        \centering
        \includegraphics[width=\linewidth]{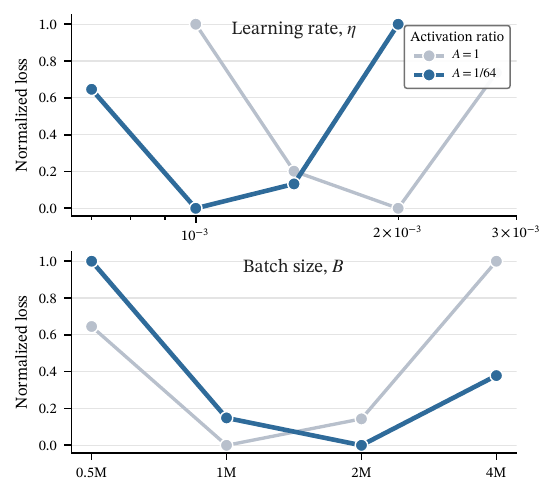}
        \caption{}
        \label{fig:intro-hyperparameter-shifts}
    \end{subfigure}
    \hfill
    \begin{subfigure}[b]{0.64\textwidth}
        \centering
        \includegraphics[width=\linewidth]{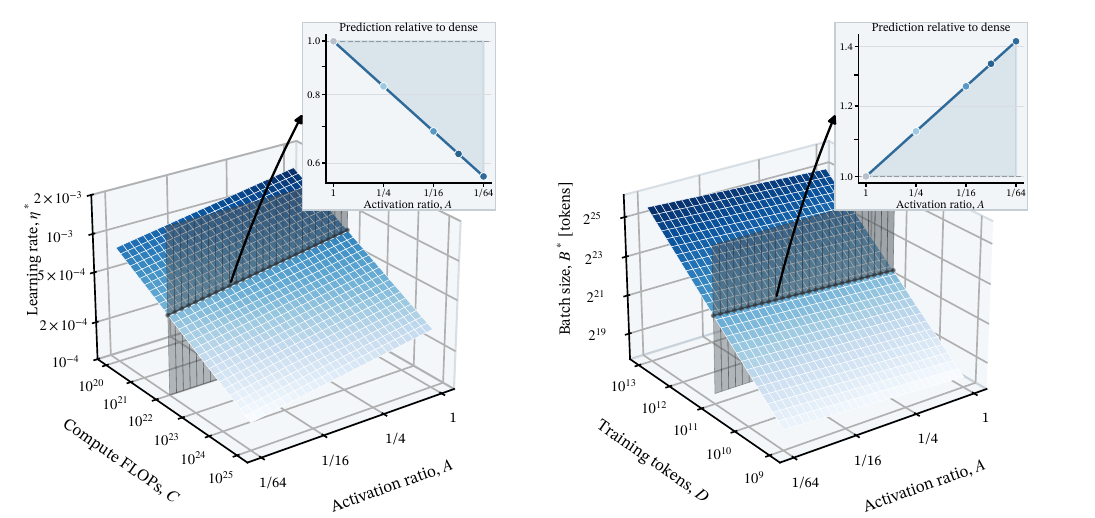}
        \caption{}
        \label{fig:intro-sparsity-aware-laws}
    \end{subfigure}
    \caption{Sparsity-dependent hyperparameter shifts and our unified hyperparameter scaling laws for MoE. (a) At 80M activated parameters and 20B training tokens, the dense model and the $A=1/64$ MoE prefer different learning rates and batch sizes. (b) Our laws predict the optimal learning rate from $(C,A)$ and batch size from $(D,A)$.}
    \label{fig:intro-sparsity-motivation}
\end{figure*}

\section{Introduction}

Mixture-of-Experts (MoE) models~\citep{shazeer2017outrageously,team2026kimi,xu2026deepseek} have emerged as an important paradigm for scaling the capacity of large language models (LLMs), activating only a small subset of experts per token to expand the total parameter count without a proportional increase in training compute~\citep{clark2022unified,tian2026towards}. As model sizes and training budgets continue to grow, hyperparameters such as the learning rate (LR) and batch size (BS) become increasingly important to training stability, convergence speed, and final performance~\citep{mccandlish2018empirical,bjorck2025scaling,zhang2025does}. Exhaustively tuning these hyperparameters at the target scale, however, is prohibitively expensive. Prior studies reduce this cost by establishing empirical scaling laws that relate the optimal learning rate and batch size to model size, dataset size, or compute budget~\citep{kaplan2020scaling,hoffmann2022training,bi2024deepseek}. Recent work has extended these laws to MoEs, but largely under conventional or fixed sparsity configurations~\citep{ludziejewski2025joint,li2025predictable,wang2024scaling,tian2026towards}.

However, prior studies report conflicting findings on MoE hyperparameters. Some find that hyperparameters transfer robustly between dense models and sparse MoEs~\citep{wang2024scaling,li2025predictable}, whereas others observe that MoEs favor larger batch sizes and lower learning rates~\citep{ludziejewski2025joint,tian2026towards}. Existing scaling laws do not characterize how the optima vary continuously with sparsity, particularly in the ultra-sparse regime down to $A=1/64$. Our experiments further show that conventional hyperparameter scaling laws are insufficient for ultra-sparse MoEs. As shown in the controlled experiments in Figure~\ref{fig:intro-sparsity-motivation}, the optimal learning rate and batch size shift between the dense model and the $A=1/64$ MoE even when activated parameter count and training data are matched. This shift indicates that the activation ratio provides additional predictive information and should be explicitly incorporated into scaling laws.

To address this gap, we conduct \textbf{1,800 pre-training runs}, systematically varying the activated non-embedding parameter count $N$, total non-embedding parameter count $N_{\mathrm{tot}}$, activation ratio $A$, and number of training tokens $D$. The experiments span six activated-parameter scales from approximately 10M to 324M and reach 6B total non-embedding parameters, processing approximately 20 trillion tokens at a cost of \textbf{200,000 equivalent H800 GPU-hours}. For each configuration, we analytically compute the non-embedding FLOPs per token $M$ and define training compute as $C=MD$. At fixed sparsity, the optimal batch size and learning rate follow power laws in $D$ and $C$, respectively. Across sparsity levels, $A$ modifies their prefactors through a multiplicative power law. These findings yield the unified form $h^*(X,A)=k_hX^{\gamma_h}A^{\delta_h}$, which we compare with alternative forms using large-scale experimental data. On a frozen target with 324M activated parameters and 12B total parameters, the law jointly extrapolates beyond the fitting ranges in $A$, $D$, and $C$, outperforming existing scaling laws~\citep{bi2024deepseek, li2025predictable} and predicting the hyperparameters closest to the observed optima.
Overall, our main contributions are as follows:
\begin{itemize}[leftmargin=*]
\item We reconcile conflicting findings in prior work within a unified experimental framework. At fixed sparsity, the optimal batch size and learning rate follow power laws in training tokens and training compute, respectively. Across sparsity levels, activation ratio provides an additional key predictor by multiplicatively modifying the prefactors of both laws.
\item We derive unified hyperparameter scaling laws across MoE sparsity levels and validate their fit on large-scale experimental data through comparisons with alternative forms. The laws jointly extrapolate to a target beyond the ranges of the development data, predict hyperparameters close to the observed optima, and transfer robustly across the tested expert granularities.
\end{itemize}

\section{Preliminaries}
\label{sec:preliminaries}

We first formulate optimal hyperparameter selection across MoE sparsity levels and then describe the controlled experimental setup used to isolate the effects of scale and sparsity.

\subsection{Problem Formulation}
\label{sec:problem-formulation}

We formalize optimal hyperparameter selection for MoEs across sparsity levels. Let $N$ denote the number of non-embedding parameters activated per token, $N_{\mathrm{tot}}$ the total number of non-embedding parameters, $D$ the number of training tokens, and $M$ the analytically computed non-embedding FLOPs per token for each architecture. We further denote the number of experts activated per token and the total number of experts by $E_{\mathrm{act}}$ and $E_{\mathrm{tot}}$, respectively, and define
\begin{equation}
    A \equiv \frac{E_{\mathrm{act}}}{E_{\mathrm{tot}}},
    \qquad
    C \equiv MD,
    \label{eq:moe-scale-definitions}
\end{equation}
where a smaller $A$ indicates a sparser model. Following previous studies~\citep{deepseekai2024deepseekv3technicalreport,tian2026towards}, we compute training compute as $C=MD$ rather than using the approximation $C\approx6ND$~\citep{kaplan2020scaling}. Thus, $C$ denotes the analytically computed non-embedding training FLOPs, while $N$ remains a parameter-scale descriptor and candidate predictive variable.

Let $\eta$ denote the peak learning rate under a fixed schedule and $B$ the global number of tokens processed per optimizer update. Holding the architecture family, data distribution, and all remaining training choices fixed, we write the validation cross-entropy after training on $D$ tokens as
\begin{equation}
    \mathcal{L}\!\left(\eta,B
    \mid N,N_{\mathrm{tot}},M,D,A\right),
    \label{eq:moe-hyperparameter-loss}
\end{equation}
and define the optimal learning rate and batch size over the candidate search space as
\begin{equation}
    (\eta^{*},B^{*})
    \equiv
    \operatorname*{arg\,min}_{\eta,\,B}
    \mathcal{L}\!\left(\eta,B
    \mid N,N_{\mathrm{tot}},M,D,A\right).
    \label{eq:optimal-moe-hyperparameters}
\end{equation}
Equation~\eqref{eq:optimal-moe-hyperparameters} defines the conceptual joint optimum over the two-dimensional LR--BS space. Because a finite grid does not directly reveal the continuous optimum, subsequent quantitative analyses consider both the observed optimum and the near-optimal set to reduce the effect of noise. Following prior work~\citep{bi2024deepseek}, we prespecify a default threshold of 0.1\% and regard the hyperparameters of model configurations whose generalization error is no more than this threshold above the minimum as near-optimal. This formulation extends standard hyperparameter-scaling setups~\citep{li2025predictable,ludziejewski2025joint,zhou2026set} by making MoE sparsity explicit. It does not assume in advance which scale variables best explain variation in $(\eta^*,B^*)$; we compare the predictive power of the candidate variables empirically in Section~\ref{sec:hyperparameter-scaling-laws}.

Following the power-law assumption adopted in prior hyperparameter-scaling studies~\citep{bi2024deepseek,li2025predictable,ludziejewski2025joint,tian2026towards}, we model the relation between an optimal hyperparameter $h^*$ and its corresponding scale variable $X$ as
\begin{equation}
    h^*(X)=aX^b,
    \label{eq:generic-hyperparameter-power-law}
\end{equation}
where $a$ is the prefactor and $b$ is the scaling exponent. Equivalently, $\log h^*=\log a+b\log X$, where $\log a$ is the intercept and $b$ is the slope in log-log space.

\subsection{Controlled Experimental Setup}
\label{sec:controlled-experimental-setup}

{To isolate the effects of model scale, training horizon, and sparsity, we construct six model scales and systematically vary $N$, $N_{\mathrm{tot}}$, $D$, and $A\in\{1,1/4,1/16,1/32\}$. The main scaling sweep uses the same mixed pre-training data, 4,096-token sequences, a hybrid linear-attention/MLA backbone~\citep{qin2023transnormerllm,deepseekai2024deepseekv3technicalreport,li2026ling}, and the Muon optimizer~\citep{liu2025muon}. Further details are provided in Appendix~\ref{sec:additional-experimental-details}.}

{For each experimental group, we search the peak learning rate $\eta$ and global token batch size $B$. Here, $B$ denotes the number of tokens processed per optimizer update. All fitting scales and activation ratios share the main grid listed in Table~\ref{tab:model-scale-grid}, while the final held-out configuration uses a separate search grid. Training follows a warmup--stable--decay (WSD) schedule~\citep{hu2024minicpm}: after a 1\% warmup, the learning rate remains at its peak before a final 10\% exponential decay. The complete model grid, compute budgets, and implementation controls are reported in Section~\ref{sec:model-configurations}.}

\section{Optimal Hyperparameters Scaling Laws for Ultra-Sparse MoEs}
\label{sec:hyperparameter-scaling-laws}

In this section, we first evaluate activation ratio as a distinct predictive dimension for the optimal learning rate and batch size in MoEs. We then separate its predictive contribution from those of data and compute and derive unified scaling laws.

\subsection{Beyond Parameter Counts: Sparsity Matters}
\label{sec:governing-scale-variables}

\begin{figure*}[t]
    \centering
    \captionsetup{skip=5pt}
    \captionsetup[subfigure]{skip=1pt}
    \begin{subfigure}[b]{0.49\textwidth}
        \centering
        \includegraphics[width=\linewidth]{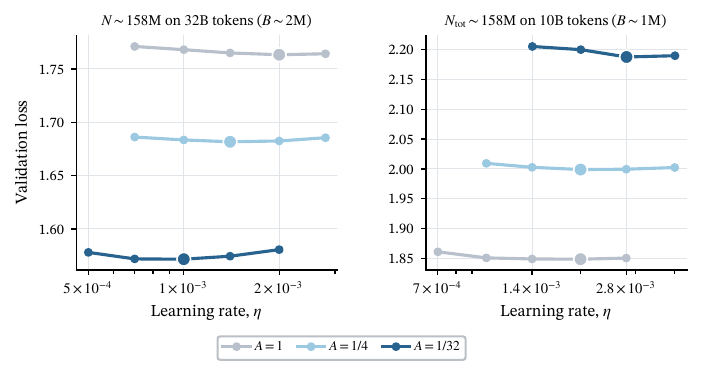}
        \caption{}
        \label{fig:parameter-count-learning-rate}
    \end{subfigure}
    \hfill
    \begin{subfigure}[b]{0.49\textwidth}
        \centering
        \includegraphics[width=\linewidth]{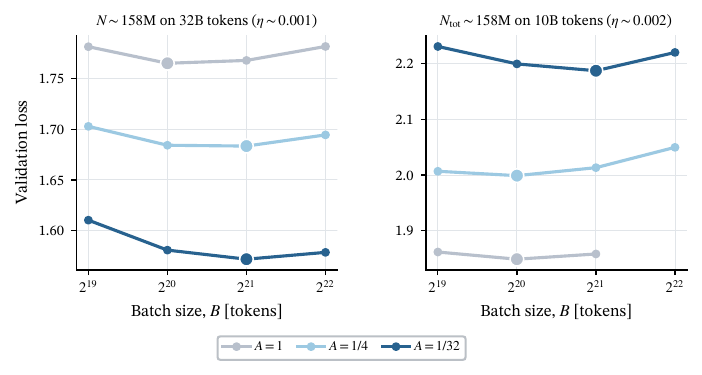}
        \caption{}
        \label{fig:parameter-count-batch-size}
    \end{subfigure}
    \caption{Optimal hyperparameters still shift with sparsity at matched activated or total parameter count. Validation loss is shown against (a) learning rate $\eta$ at a fixed batch size and (b) batch size $B$ at a fixed learning rate. Within each subfigure, the left panel matches activated parameter count $N$, while the right panel matches total parameter count $N_{\mathrm{tot}}$. Color denotes activation ratio $A$. Larger markers show the minimum along each plotted curve.}
    \label{fig:parameter-count-insufficiency}
\end{figure*}

Existing studies agree that model scale affects the optimal learning rate, but disagree on whether the relevant scale is the activated parameter count~\citep{ludziejewski2025joint} or the total parameter count~\citep{li2025predictable}. We first fix the batch size and test whether alignment by either parameter count removes the learning-rate shifts across sparsity levels. As shown in Figure~\ref{fig:parameter-count-learning-rate}, the curve minima still shift across activation ratios at matched activated parameter count $N$ (left). Alignment by total parameter count $N_{\mathrm{tot}}$ likewise fails to remove the shift (right). Thus, neither parameter count alone explains the variation in the optimal learning rate.

The appropriate scale variable for predicting the optimal batch size is similarly unsettled: some studies model it as a function of compute $C$~\citep{bi2024deepseek,team2025every}, while others tie it primarily to the number of training tokens $D$~\citep{li2025predictable}. Figure~\ref{fig:parameter-count-batch-size} shows that, at a fixed learning rate, neither variable provides a unified explanation of the optimal batch size across sparsity levels. Even when $D$, $N$, and total training compute $C=MD$ are held fixed, $B^*$ still varies with $A$ (left). Thus, neither training tokens nor compute alone explains the optimal batch size. Alignment by total parameter count $N_{\mathrm{tot}}$ leads to the same conclusion (right).

These results provide practical guidance for selecting candidate variables: the effect of sparsity is not absorbed by compute, activated parameter count, or total parameter count. We therefore use the activation ratio $A$ to represent sparsity explicitly, distinguishing models with similar per-token compute but different expert capacities and supplying the predictive dimension needed for scaling across sparsity levels. We evaluate the contribution of $A$ through grouped out-of-fold prediction on the observed two-dimensional surfaces.

\begin{tcolorbox}[
    colback=gray!15,
    colframe=gray!45,
    boxrule=0.6pt,
    arc=2pt,
    left=6pt,
    right=6pt,
    top=5pt,
    bottom=5pt
]
\textbf{Takeaway: Sparsity introduces an additional scaling dimension for optimal hyperparameters.} Parameter counts, training tokens, and compute alone cannot explain the shifts in $(\eta^*,B^*)$ across sparsity levels; the activation ratio $A$ must therefore be modeled explicitly.
\end{tcolorbox}

\subsection{Disentangling Compute, Data, and Sparsity Effects}
\label{sec:univariate-hyperparameter-scaling}

The preceding subsection shows that the effect of activation ratio on optimal hyperparameters must be modeled explicitly. To separate this effect from the base scale dependencies, we first hold activation ratio fixed and use the estimated optima extracted from the observed two-dimensional loss surfaces to compare which scale variables best explain the optimal learning rate and batch size. We then model the additional effect of activation ratio separately.

\begin{figure*}[t]
    \centering
    \captionsetup{skip=5pt}
    \captionsetup[subfigure]{skip=1pt}
    \begin{subfigure}[b]{0.64\textwidth}
        \centering
        \includegraphics[width=\linewidth]{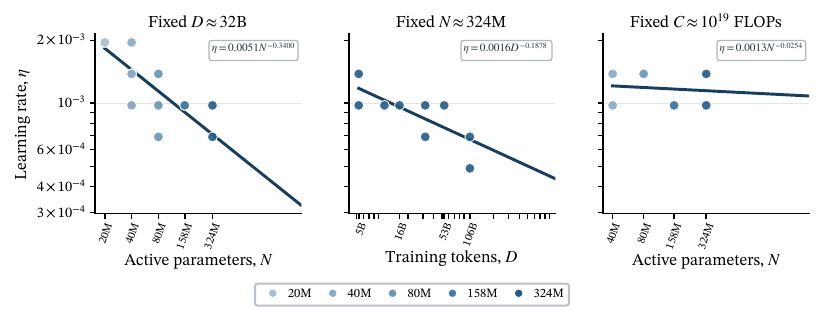}
        \caption{}
        \label{fig:lr-variable-comparison}
    \end{subfigure}
    \hfill
    \begin{subfigure}[b]{0.35\textwidth}
        \centering
        \includegraphics[width=\linewidth]{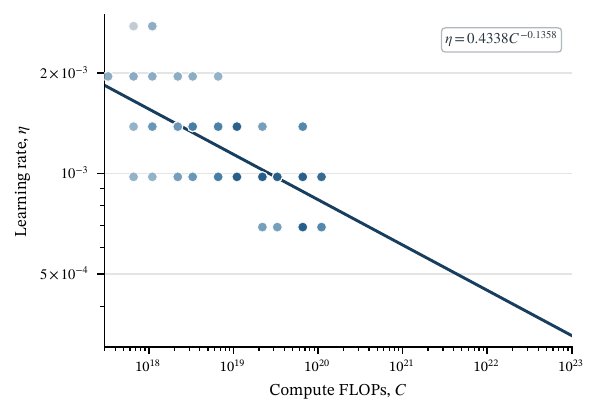}
        \caption{}
        \label{fig:lr-compute-power-law}
    \end{subfigure}
    \caption{Training compute organizes the optimal learning rate at fixed sparsity. (a) Comparison of $N$, $D$, and $C$ as candidate predictive variables for $\eta^*$. (b) Power-law relation between $\eta^*$ and $C$ on the representative $A=1/32$ slice.}
    \label{fig:lr-dependencies}
\end{figure*}
\begin{figure*}[t]
    \centering
    \captionsetup{skip=5pt}
    \captionsetup[subfigure]{skip=1pt}
    \begin{subfigure}[b]{0.64\textwidth}
        \centering
        \includegraphics[width=\linewidth]{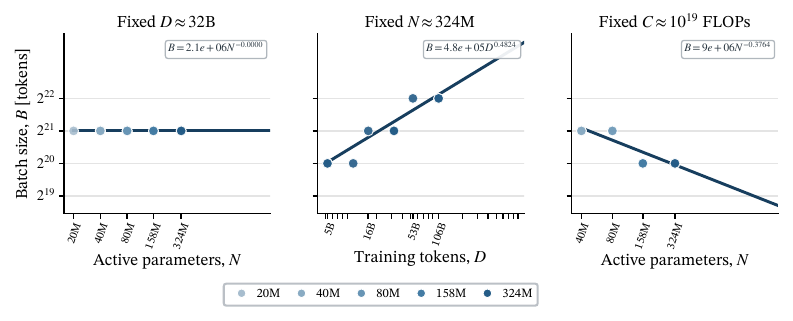}
        \caption{}
        \label{fig:bs-variable-comparison}
    \end{subfigure}
    \hfill
    \begin{subfigure}[b]{0.335\textwidth}
        \centering
        \includegraphics[width=\linewidth]{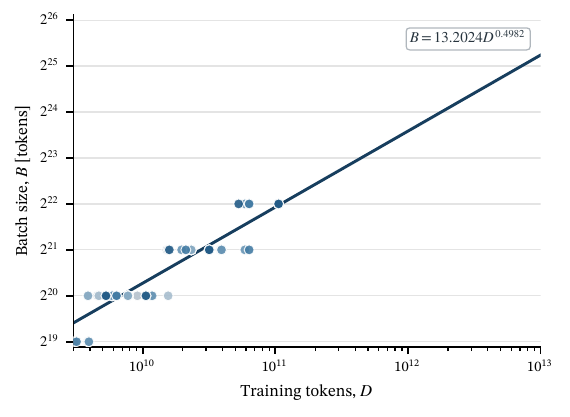}
        \caption{}
        \label{fig:bs-data-power-law}
    \end{subfigure}
    \caption{Training tokens organize the optimal batch size at fixed sparsity. (a) Comparison of $N$, $C$, and $D$ as candidate predictive variables for $B^*$. (b) Power-law relation between $B^*$ and $D$ on the representative $A=1/32$ slice.}
    \label{fig:bs-dependencies}
\end{figure*}

\paragraph*{Compute Scaling of Optimal Learning Rate}
\label{sec:compute-scaling-learning-rate}

At fixed sparsity, existing studies make different assumptions about the base scale of the optimal learning rate: some model it jointly with $N$ and $D$~\citep{bjorck2025scaling,li2025predictable}, whereas others use training compute $C$ directly~\citep{bi2024deepseek,team2025every}. Under our definition $C=MD$, the key question is whether $\eta^*$ remains sensitive to the allocation between per-token compute $M$ and training duration $D$ at fixed $C$. We retain $N$ as a parameter-count baseline, compare $N$, $D$, and $C$ as explanatory variables, and vary the $M/D$ allocation while holding $C$ fixed. 
Figure~\ref{fig:lr-variable-comparison} shows that neither $N$ nor $D$ alone organizes the optimal learning rates across configurations, whereas $C$ provides a clearer relation. Moreover, at the same $C$, the optimal learning rate remains nearly unchanged despite large differences in the allocation of $M$ and $D$. As shown in Figure~\ref{fig:lr-compute-power-law}, $\eta^*$ follows a stable power law in $C$ on the representative $A=1/32$ slice. Using $C$ in place of the separate variables $N$ and $D$ reduces the model degrees of freedom, making the scaling relation more stable and easier to fit.

\paragraph*{Data Scaling of Optimal Batch Size}
\label{sec:data-scaling-batch-size}

At fixed sparsity, the base scale for the optimal batch size is likewise disputed: some studies use training compute $C$~\citep{bi2024deepseek,team2025every}, while others identify the number of training tokens $D$ as the primary variable~\citep{li2025predictable,bergsma2026power}. We apply the same controlled comparison used for the learning rate to evaluate whether $N$, $C$, or $D$ most consistently organizes the observed batch-size optima. 
Figure~\ref{fig:bs-variable-comparison} shows that $B^*$ is organized primarily by $D$, while neither $N$ nor $C$ yields a comparably stable relation. On the representative $A=1/32$ slice, $B^*$ follows a stable power law in $D$, as shown in Figure~\ref{fig:bs-data-power-law}. Thus, longer training horizons favor larger global token batches, consistent with prior observations~\citep{li2025predictable,bergsma2026power}.

\paragraph*{Sparsity as an Additional Scaling Dimension}
\label{sec:sparsity-additional-dimension}

\begin{figure}[t]
    \centering
    \includegraphics[width=\linewidth]{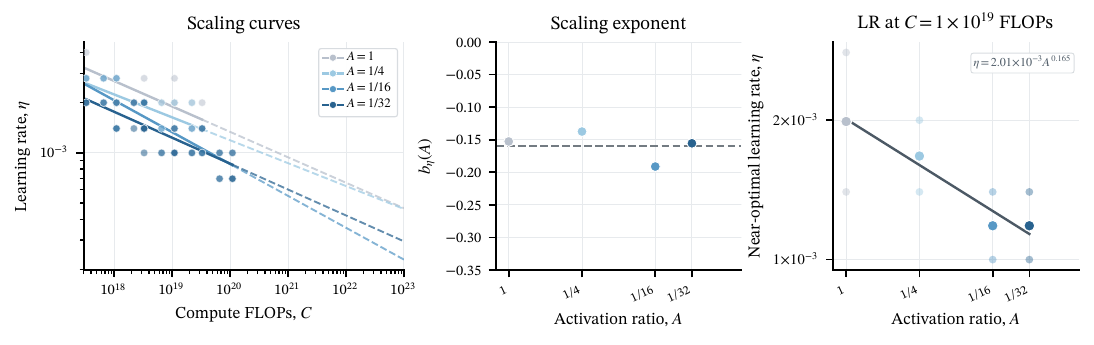}
    \caption{Sparsity-dependent shifts in the optimal-learning-rate power law. Left: near-optimal $\eta^*$ and power-law fits versus $C$ for four activation ratios; dashed extensions indicate extrapolation beyond each observed range. Middle: fitted exponent $b_\eta(A)$ versus $A$; the horizontal dashed line marks the mean, and the similar exponents motivate a shared-exponent model. Right: near-optimal learning rates at matched FLOPs. Small markers show individual observations, large markers show geometric means, and the solid line shows a multiplicative power-law fit in $A$.}
    \label{fig:sparsity-dependencies-lr}
\end{figure}
\begin{figure}[t]
    \centering
    \includegraphics[width=\linewidth]{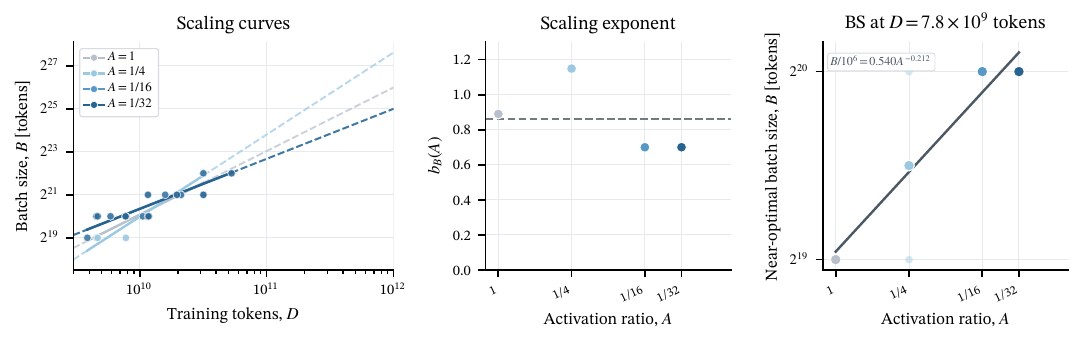}
    \caption{Sparsity-dependent shifts in the optimal-batch-size power law. Left: near-optimal $B^*$ and power-law fits versus $D$ for four activation ratios; dashed extensions indicate extrapolation beyond each observed range. Middle: fitted exponent $b_B(A)$ versus $A$; the horizontal dashed line marks the mean, and the similar exponents motivate a shared-exponent model. Right: near-optimal batch sizes at matched training-token counts. Small markers show individual observations, large markers show geometric means, and the solid line shows a multiplicative power-law fit in $A$.}
    \label{fig:sparsity-dependencies-bs}
\end{figure}

To characterize how sparsity modifies these relations, we fit a separate scaling curve at each observed slice $A\in\{1,1/4,1/16,1/32\}$, compare the fitted coefficients, and analyze how the estimated optimal hyperparameters vary with $A$ at fixed $C$ or $D$.

Figure~\ref{fig:sparsity-dependencies-lr} summarizes the fitted coefficients across activation ratios. The learning-rate exponents $b_\eta(A)$ fluctuate around their mean, while at fixed $C$, $\log\eta^*$ is approximately linear in $\log_2 A$, equivalently $\eta^*\propto A^{\delta_\eta}$. Appendix Figure~\ref{fig:appendix-lr-sparsity-slices} shows the corresponding relationship across three additional fixed-compute slices. These descriptive results suggest that sparsity may act primarily through a multiplicative prefactor correction. Because a smaller $A$ yields a lower optimal learning rate, the candidate multiplicative model has $\delta_\eta>0$. 
Batch size exhibits the same pattern but shifts in the opposite direction. In Figure~\ref{fig:sparsity-dependencies-bs}, the fitted exponents $b_B(A)$ likewise fluctuate around their mean, while at fixed $D$, $\log B^*$ is approximately linear in $\log_2 A$, equivalently $B^*\propto A^{\delta_B}$. Because a smaller $A$ yields a larger optimal batch size, the candidate multiplicative model has $\delta_B<0$. Appendix Figure~\ref{fig:appendix-bs-sparsity-slices} shows the corresponding relationship across three fixed-token slices. This trend is qualitatively consistent with prior observations~\citep{team2025every,ludziejewski2025joint}.

Our theoretical analysis shows that a simple gradient-noise model can explain both trends. Under balanced routing, each expert receives only about $AB$ tokens per step, so decreasing $A$ increases expert-side gradient noise and favors a larger global batch. Because the optimal batch size does not grow enough to keep the effective expert batch $AB^*$ constant, the optimal learning rate still decreases as $A$ decreases. Appendix~\ref{sec:appendix-theoretical-analysis} provides the full derivation.

Overall, the results show that $C$ and $D$ organize the base power laws for the optimal learning rate and batch size, respectively, while the observed shifts along $A$ support the separable correction $A^{\delta_h}$ as a candidate form. Because four slices cannot rule out an interaction in which the exponent varies with $A$, Section~\ref{sec:scaling-law-validation} compares the shared-exponent and interaction forms using identical grouped folds rather than selecting a model from the slice plots alone.

\begin{tcolorbox}[
    colback=gray!8,
    colframe=gray!45,
    boxrule=0.6pt,
    arc=2pt,
    left=6pt,
    right=6pt,
    top=5pt,
    bottom=5pt
]
\noindent\textbf{Takeaway: Training compute $C$, training tokens $D$, and activation ratio $A$ provide three complementary dimensions for predicting the optimal hyperparameters of MoEs.}

\par\smallskip

\begin{itemize}[leftmargin=*,nosep]
\item \textbf{Base scaling at fixed sparsity.} $B^*$ follows a power law in training tokens $D$, while $\eta^*$ follows a power law in compute $C=MD$ and is insensitive to the $M/D$ allocation at fixed $C$.

\item \textbf{Multiplicative sparsity correction.} The observed shifts along activation ratio $A$ support the separable factor $A^{\delta_h}$ as a sparsity correction.
\end{itemize}
\end{tcolorbox}

\subsection{Unified Hyperparameter Scaling Laws for MoEs}
\label{sec:joint-hyperparameter-scaling-laws}

\paragraph*{Empirical Structure}

To characterize the relation between activation ratio and the base scale variables, we summarize three empirical observations from the preceding analysis:
\begin{enumerate}[leftmargin=*,nosep]
\item At fixed $A$, $\eta^*$ follows a power law in $C$, while $B^*$ follows a power law in $D$. Thus, $C$ and $D$ serve as their respective base predictive variables.

\item Across activation ratios, the fitted exponents fluctuate around their respective means, supporting a shared-exponent candidate.

\item At fixed $C$, $\log\eta^*$ is approximately linear in $\log_2 A$, with $\delta_\eta>0$; at fixed $D$, $\log B^*$ is approximately linear in $\log_2 A$, with $\delta_B<0$. Thus, the sparsity effect on each hyperparameter can be represented by a multiplicative power law in $A$.
\end{enumerate}

Observation 1 supports $C$ and $D$ as the base predictive variables for learning rate and batch size, respectively. Observations 2 and 3 support sharing the base scaling exponent across activation ratios and applying a multiplicative correction in $A$ to the prefactor. Based on these observations, we propose a unified hyperparameter formulation.

\paragraph*{Unified Hyperparameter Formulation}
\label{sec:joint-hyperparameter-formulation}

We express both laws with the common functional family
\begin{equation}
    h^*(X,A)
    = k_h X^{\gamma_h} A^{\delta_h},
    \qquad
    (h,X)\in\{(\eta,C),(B,D)\}.
    \label{eq:symbolic-hyperparameter-law}
\end{equation}
Here, $k_h$ is a constant prefactor independent of scale and sparsity, $\gamma_h$ is the scaling exponent for the base variable $X$, and $\delta_h$ is the sparsity exponent for the activation ratio $A$. In log space,
\begin{equation}
    \log h^*=\log k_h+\gamma_h\log X+\delta_h\log A.
    \label{eq:log-symbolic-hyperparameter-law}
\end{equation}
Specifically, learning rate and batch size use training compute $C$ and training-token count $D$ as their base scale variables, with multiplicative prefactor corrections in $A$, where $\delta_\eta>0$ and $\delta_B<0$:
\begin{align}
    \eta^*(C,A) &= k_\eta C^{\gamma_\eta} A^{\delta_\eta},
    \label{eq:symbolic-learning-rate-law}\\
    B^*(D,A) &= k_B D^{\gamma_B} A^{\delta_B}.
    \label{eq:symbolic-batch-size-law}
\end{align}
These formulations clearly separate the base power laws at fixed sparsity from the multiplicative sparsity corrections. The exponents $\gamma_\eta$ and $\gamma_B$ describe scaling with compute and training tokens, respectively, while $\delta_\eta$ and $\delta_B$ measure sensitivity to the activation ratio. The shared-exponent assumption is motivated by the preceding slice trends rather than established by the plots alone. Subsequent analysis compares it with a more flexible interaction form and uses grouped out-of-fold prediction to determine which functional form is better supported by the experimental results.

\paragraph*{Comparison with Existing Scaling Laws}
\label{sec:comparison-existing-scaling-laws}

Table~\ref{tab:hyperparameter-scaling-variable-comparison} compares representative scaling laws in terms of the scale variables used for the optimal learning rate and batch size. 
Unlike existing formulations, ours uses $C$ and $D$ as the respective base predictive variables for learning rate and batch size, and uses the activation ratio $A$ to explicitly describe continuous shifts across sparsity levels.

\begin{table}[H]
    \centering
    \caption{Comparison of scale variables used to model optimal learning rates and batch sizes.}
    \small
    \label{tab:hyperparameter-scaling-variable-comparison}
    \begin{tabular}{@{}p{0.52\linewidth}>{\centering\arraybackslash}p{0.20\linewidth}>{\centering\arraybackslash}p{0.20\linewidth}@{}}
        \toprule
        Method & Variables for $\eta^*$ & Variables for $B^*$ \\
        \midrule
        DeepSeek Law~\citep{bi2024deepseek}
        & $C$ & $C$ \\
        Microsoft Law~\citep{bjorck2025scaling}
        & $\bigl(N_{\text{tot}}, D\bigr)$ & -- \\
        Joint MoE Scaling Law~\citep{ludziejewski2025joint}
        & $\bigl(N, E_{\text{tot}}\bigr)$
        & -- \\
        Step Law~\citep{li2025predictable}
        & $\bigl(N_{\text{tot}}, D\bigr)$ & $D$ \\
        \midrule
        Ours
        & $\bigl(C,A\bigr)$ & $\bigl(D,A\bigr)$ \\
        \bottomrule
    \end{tabular}
\end{table}

\section{Fitting and Predictive Validation}
\label{sec:scaling-law-validation}

Next, we use large-scale experimental data to estimate the coefficients of the unified scaling laws developed above. We then evaluate their fit quality and robustness through grouped out-of-fold prediction, and finally perform a single-point consistency check for joint extrapolation in $A$, $D$, and $C$ on a held-out target with 12B total parameters and an activation ratio of $A=1/64$.

\subsection{Fitting Protocol and Fitted Laws}
\label{sec:fitting-protocol}

The formal analysis uses only development data at $A\in\{1,1/4,1/16,1/32\}$ across the six activated-parameter scales. Functional-family selection, grouped cross-validation, and final coefficient fitting are all restricted to this set. Throughout fitting, cross-validation, fixed-$C$ slicing, and held-out evaluation, compute is obtained from the configuration-specific analytical value as $C=MD$. Following prior work, we prespecify a default threshold of $0.1\%$ and define the near-optimal set for each LR--BS loss surface as the observed grid points whose losses are no more than this threshold above the observed minimum, reducing sensitivity to noise in any single grid optimum. We transform each power law into a linear form in log space and fit its parameters by least squares. Appendix~\ref{sec:model-configurations} provides the complete experimental design.

\begin{table}[H]
    \centering
    \caption{Fitted coefficients of the unified scaling laws defined in Equations~\eqref{eq:symbolic-learning-rate-law} and~\eqref{eq:symbolic-batch-size-law}.}
    \label{tab:fitted-law-coefficients}
    \small
    \begin{tabular}{@{}lcccc@{}}
        \toprule
        Hyperparameter $h$ & Input variables & $k_h$ & $\gamma_h$ & $\delta_h$ \\
        \midrule
        Learning rate $\eta$ & $(C,A)$ & $0.8343$ & $-0.1385$ & $0.1361$ \\
        Batch size $B$ & $(D,A)$ & $6.4765$ & $0.5181$ & $-0.0841$ \\
        \bottomrule
    \end{tabular}
\end{table}

Table~\ref{tab:fitted-law-coefficients} summarizes the fitted coefficients of the two laws. Here, $C=MD$ is measured in non-embedding training FLOPs, while $D$ and $B$ are measured in tokens; the coefficient values therefore depend on these units. The multiplicative sparsity exponents $\delta_\eta>0$ and $\delta_B<0$ show that decreasing $A$ lowers the optimal learning rate and increases the optimal batch size.

\subsection{Fit Quality and Robustness}

To evaluate the fit quality and robustness of our scaling laws, we assess out-of-fold predictions using two grouped cross-validation schemes:
\begin{itemize}
    \item \textbf{Leave-one-activation-ratio-out (LOAO)} holds out one of the four  activation ratios and removes all LR--BS loss surfaces at that ratio across active-parameter scales. The optima are then re-extracted from the remaining activation ratios, and all coefficients are refitted.
    \item \textbf{Leave-one-active-scale-out (LONO)} holds out one of the six active-parameter scales and removes all LR--BS loss surfaces at that scale across activation ratios. The optima are then re-extracted from the remaining active-parameter scales, and all coefficients are refitted.
\end{itemize}
The purpose of LOAO and LONO is to evaluate which candidate functional family best captures the scaling of the estimated optimal hyperparameter coordinates, rather than to predict validation loss. We therefore use the absolute log-ratio errors in LR and BS as the primary metrics, while practical loss differences are evaluated separately on the final held-out target in Section~\ref{sec:heldout-extrapolation}. To prevent information leakage, we construct each fold at the loss-surface level and use only the training split to extract optima and estimate coefficients. For each hyperparameter $h$, let $\mathcal G$ denote the set of held-out groups and $s$ index the eligible loss surfaces. We define the overall error as the equally weighted mean of the median absolute $\log_2$-ratio error within each group:
\begin{equation}
    e_h=\frac{1}{|\mathcal G|}\sum_{g\in\mathcal G}
    \operatorname{median}_{s\in g}\left|\log_2(\hat h_s/h_s^*)\right|,
    \label{eq:grouped-prediction-error}
\end{equation}

\begin{figure*}[t]
    \centering
    \includegraphics[width=\textwidth]{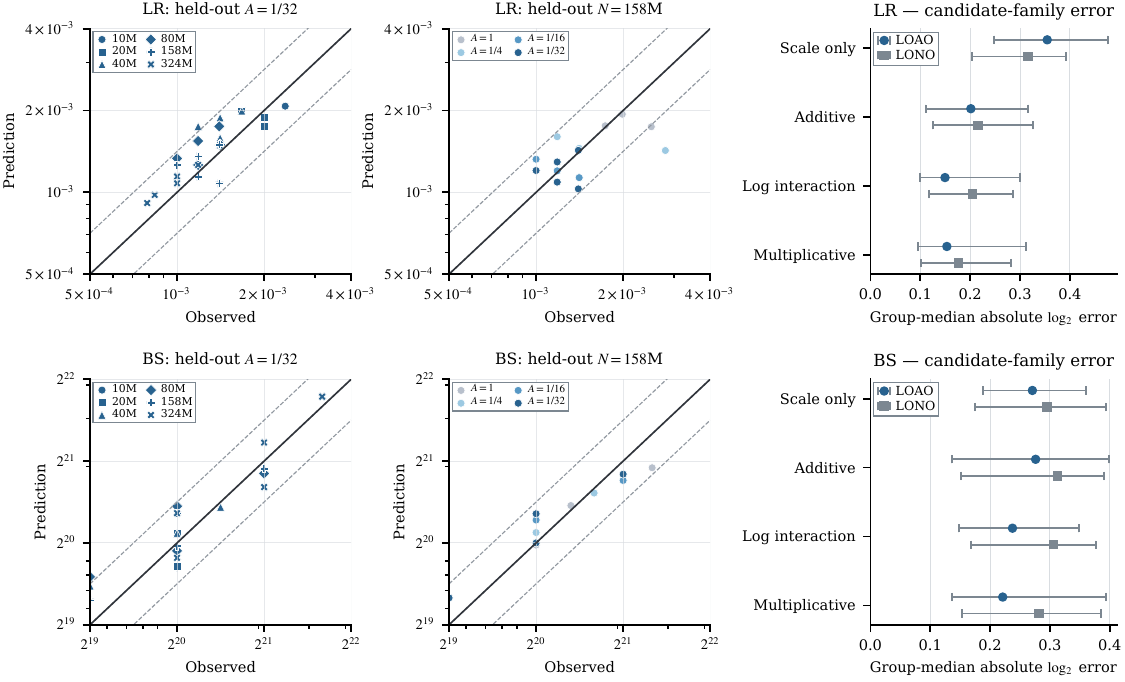}
    \caption{Grouped prediction results for the multiplicative scaling laws. Each point represents an LR--BS loss surface. The left and middle columns show LOAO predictions with $A=1/32$ held out and LONO predictions with the 158M scale held out, respectively. Solid lines denote equality, and dashed lines mark factor-$2^{0.5}$ error bounds. The right column compares LOAO and LONO errors with conditional 95\% paired surface-cluster bootstrap intervals.}
    \label{fig:joint-law-validation}
\end{figure*}

\begin{table*}[b]
    \centering
    \caption{Hyperparameter prediction errors for the four candidate families. Each entry reports the LONO and LOAO absolute base-2 log-ratio errors, in that order.}
    \label{tab:candidate-functional-families}
    \small
    {%
    \begin{tabular}{@{}llccc@{}}
        \toprule
        Candidate family & Functional form & $p$ & BS error & LR error \\
        \midrule
        Scale only & $kX^\gamma$ & 2 & $0.295/0.271$ & $0.316/0.355$ \\
        Additive & $k_X X^\gamma+k_A A^\delta$ & 4 & $0.312/0.276$ & $0.216/0.201$ \\
        Log interaction & $kX^{\beta_X+\beta_{XA}\log_2 A}A^{\beta_A}$ & 4 & $0.307/{0.238}$ & $0.205/\mathbf{0.150}$ \\
        \midrule
        Multiplicative (Ours, selected) & $kX^\gamma A^\delta$ & 3 & $\mathbf{0.282}/\mathbf{0.221}$ & $\mathbf{0.176}/0.153$ \\
        \bottomrule
    \end{tabular}%
    }
\end{table*}

To validate the functional form in Equation~\eqref{eq:symbolic-hyperparameter-law}, Table~\ref{tab:candidate-functional-families} compares four candidate families, scale only, additive, log interaction, and multiplicative, using identical target loss surfaces and grouped folds. The log-interaction family allows the scaling exponent to vary with $A$. For LR, all three $A$-aware families outperform the scale-only law under both cross-validation schemes; for BS, the differences among candidate families are smaller and vary with the grouping scheme. The multiplicative law has the lowest point estimate on three of the four error measures, while log interaction is marginally lower for LR under LOAO ($0.150$ versus $0.153$), indicating similar predictive performance. The left and middle columns of Figure~\ref{fig:joint-law-validation} show the held-out $A=1/32$ LOAO fold and the held-out 158M LONO fold, respectively; the right column summarizes the out-of-fold errors and conditional 95\% paired surface-cluster bootstrap intervals for all four families. Given their comparable predictive performance, we select the multiplicative law as our working model because it uses one fewer parameter and offers a simpler interpretation. Fitted coefficients for the three $A$-aware candidate families are reported in Appendix~\ref{sec:appendix-extrapolation-ablation}.

These results provide initial support for the proposed functional form. The next subsection further evaluates its joint extrapolation by applying it to a target beyond the fitting ranges in $A$, $D$, and $C$, while retaining $N$ at the observed boundary scale.

\subsection{Joint Extrapolation to a Held-Out Configuration}
\label{sec:heldout-extrapolation}

\paragraph*{Held-out configuration.}
The final held-out target has $N=324\mathrm{M}$ activated parameters and $N_{\mathrm{tot}}=12\mathrm{B}$ total parameters, with $A=1/64$ and $D=159\mathrm{B}$ training tokens. Its analytical $M$ gives a training compute of $C=MD=3\times10^{20}$ FLOPs. While $N$ is the largest scale observed in the fitting set, $A$, $D$, and $C$ all lie beyond their fitting ranges. The target losses are excluded from functional-family selection, cross-validation, and coefficient fitting until the prediction coordinates are frozen. The pre-specified grid in the holdout row of Appendix Table~\ref{tab:model-scale-grid} serves as the search reference. 

\paragraph*{Baseline protocol.}
We evaluate existing scaling laws in two complementary ways. \emph{Published-coefficient transfer} directly applies the original coefficients to test transfer across optimizers, training schedules, architectures, and parameter-count definitions, but is not included in the primary functional-form ranking. \emph{Refitted-family comparison} uses a common optimum definition and grouped folds to select each family on the formal fitting set, then re-estimates its coefficients on the complete fitting set. All coefficients and exact prediction coordinates are frozen before the held-out losses are inspected. The joint exact-point comparison includes only the DeepSeek Law~\citep{bi2024deepseek} and Step Law~\citep{li2025predictable}, which predict both LR and BS; Joint MoE Scaling Laws~\citep{ludziejewski2025joint} and Microsoft Law~\citep{bjorck2025scaling} are omitted because they lack a joint BS law.

\begin{figure*}[t]
    \centering
    \captionsetup{skip=5pt}
    \captionsetup[subfigure]{skip=1pt}
    \begin{subfigure}[b]{0.49\textwidth}
        \centering
        \includegraphics[width=\linewidth]{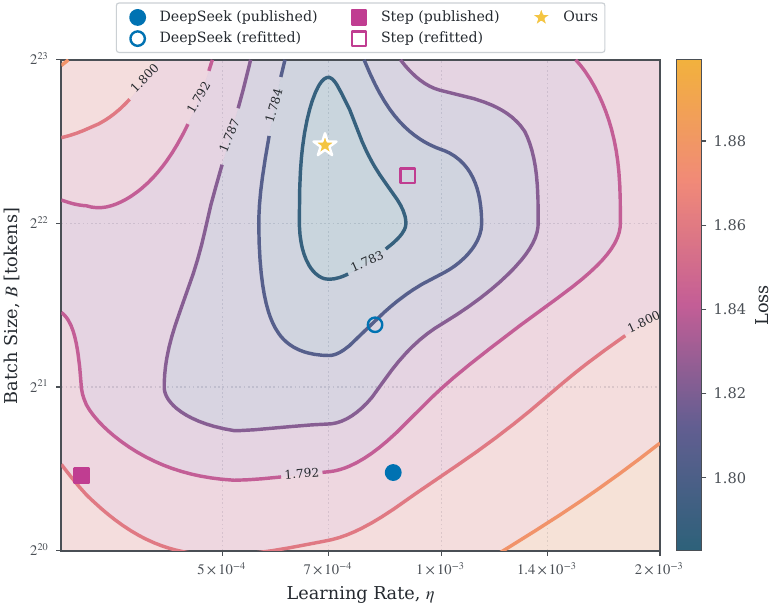}
        \caption{Training-loss search surface.}
        \label{fig:hold-out-validation}
    \end{subfigure}
    \hfill
    \begin{subfigure}[b]{0.49\textwidth}
        \centering
        \includegraphics[width=\linewidth]{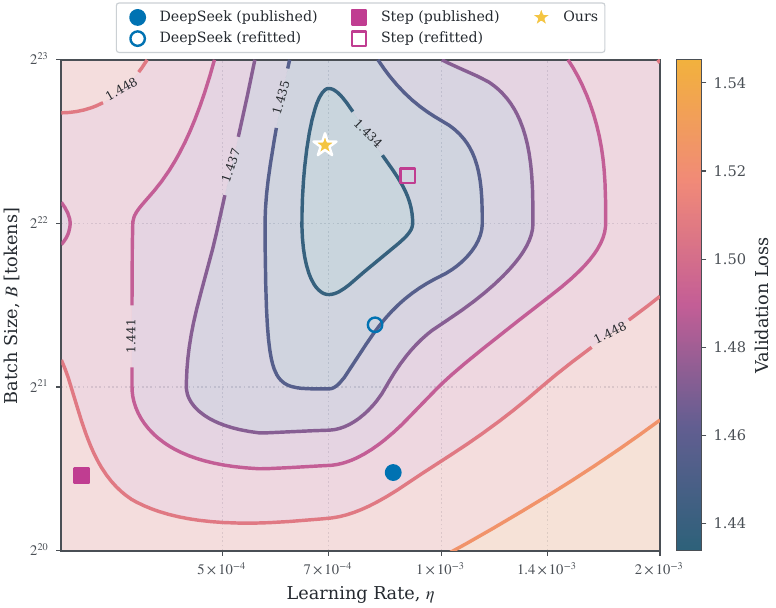}
        \caption{Validation-loss evaluation surface.}
        \label{fig:hold-out-validation-test}
    \end{subfigure}
    \caption{Training and validation loss surfaces for the held-out joint-extrapolation target. The left panel shows the training loss used for the grid search, and the right panel shows the validation loss. All predictions are frozen before inspection of the held-out losses: the star marks our prediction, and the other markers show predictions from the comparison laws.}
    \label{fig:hold-out}
\end{figure*}

\paragraph*{Held-out evaluation.}
Table~\ref{tab:heldout-12b-comparison} reports each method's exact predicted settings, frozen before inspection of the held-out losses, together with their gaps relative to our predicted setting. Figure~\ref{fig:hold-out} shows where these predictions lie on the training and validation loss surfaces. For method $m$, the relative training loss gap in per mille is defined as $1000(\widehat{\mathcal L}_m-\widehat{\mathcal L}_{\mathrm{ours}})/\widehat{\mathcal L}_{\mathrm{ours}}$. Results are reported separately for published-coefficient transfer and for refitting on the same samples used by our method. Further limitations are discussed in Section~\ref{sec:limitations}.

\begin{table*}[h]
    \centering
    \caption{Predictions on the held-out joint-extrapolation target. ``Published'' uses the original coefficients, while ``refitted'' re-estimates the same family using the same samples as our method. All methods are evaluated at $N=324\mathrm{M}$, $N_{\mathrm{tot}}=12\mathrm{B}$, $D=159\mathrm{B}$, $A=1/64$, and $C=3\times10^{20}$. Relative gaps are measured against our predicted setting.}
    \label{tab:heldout-12b-comparison}
    \small
    \setlength{\tabcolsep}{3pt}
    \begin{tabular}{@{}llp{0.36\textwidth}ccc@{}}
        \toprule
         & Mode & Formula & Predicted LR & Predicted BS
         & \makecell{Loss gap vs. ours} \\
        \midrule
        \multirow{2}{*}{DeepSeek Law} & Published
        & \makecell[l]{$\eta^*=0.3118C^{-0.1250}$\\$B^*=0.2920C^{0.3271}$}
        & $8.59\times10^{-4}$ & $1.46\times10^6$ & 7.22 \textperthousand \\
        & Refitted
        & \makecell[l]{$\eta^*=1.6763C^{-0.1619}$\\$B^*=182.9951C^{0.2038}$}
        & $8.11\times10^{-4}$ & $2.73\times10^6$ & 1.37 \textperthousand \\
        \midrule
        \multirow{2}{*}{Step Law} & Published
        & \makecell[l]{$\eta^*=1.7900N_{\mathrm{tot}}^{-0.7130}D^{0.3070}$\\$B^*=0.5800D^{0.5710}$}
        & $3.20\times10^{-4}$ & $1.44\times10^6$ & 9.02 \textperthousand \\
        & Refitted
        & \makecell[l]{$\eta^*=0.0638 N_{\mathrm{tot}}^{-0.1743}D^{-0.0084}$\\$B^*=5.2828D^{0.5345}$}
        & $8.99\times10^{-4}$ & $5.13\times10^6$ & 1.03 \textperthousand \\
        \midrule
        Ours & 
        & \makecell[l]{$\eta^*=0.8343 C^{-0.1385} A^{0.1361}$\\$B^*=6.4765 D^{0.5181} A^{-0.0841}$}
        & $6.92\times10^{-4}$ & $5.84\times10^6$ & -- \\
        \bottomrule
    \end{tabular}
\end{table*}

\subsection{Expert-Granularity Transfer and Sparsity Control}
\label{sec:expert-granularity-transfer}

We use two controlled comparisons to separate the effects of expert granularity and sparsity. The first holds the activation ratio $A$ fixed while varying the active expert count, total expert count, and expert width, testing whether the activation-ratio relation transfers across expert granularities. The second holds the total expert count and total capacity fixed while changing the active expert count and hence $A$, testing whether the resulting sparsity shift follows our scaling laws.

\paragraph*{Controlled comparisons.}
All three configurations share the backbone of the approximately 10M-activated-parameter configuration in the main sweep and are trained on $D=4.56\mathrm{B}$ tokens. Here, ``10M'' labels the backbone configuration rather than an exactly matched activated parameter count across the three controls. The reference configuration is $(E_{\mathrm{act}},E_{\mathrm{tot}},h_{\mathrm{MoE}})=(2,64,384)$, corresponding to $A=1/32$. The expert-granularity control uses $(4,128,192)$: it keeps $A=1/32$ while doubling the active and total expert counts and halving the expert width, thereby matching activated and total capacity. The sparsity control uses $(4,64,384)$: relative to the reference, it keeps the total expert count, expert width, $N_{\mathrm{tot}}$, and total routed capacity fixed while increasing $A$ to $1/16$. Apart from these MoE expert settings, all three configurations share the remaining training settings and LR--BS search grid. Complete configurations are provided in Appendix~\ref{sec:appendix-granularity-controlled-configuration}.

\paragraph*{Results.}
We first compare the left and middle panels of Figure~\ref{fig:expert-granularity-controls} to test transfer across expert granularities. The two capacity-matched configurations at fixed $A$ have similar near-optimal validation-loss regions and LR--BS optimum locations. Thus, doubling both the active and total expert counts does not materially change the optimal hyperparameters, supporting transfer of the activation-ratio relation across the tested expert granularities. We then compare the left and right panels. Holding the total expert count, expert width, $N_{\mathrm{tot}}$, and total capacity fixed while increasing $A$ from $1/32$ to $1/16$ shifts both the optimal learning rate and batch size downward, consistent with the direction predicted by our scaling laws. Together, these controls show that the sparsity effect captured by $A$ cannot be attributed solely to any single absolute expert count or to total capacity.

\providecommand{\controlledheatmap}[1]{%
    \IfFileExists{#1}{%
        \includegraphics[width=\linewidth]{#1}%
    }{%
        \begingroup
        \setlength{\fboxsep}{0pt}%
        \fcolorbox{blue}{blue!5}{%
            \begin{minipage}[c][0.56\linewidth][c]{0.97\linewidth}
                \centering\color{blue}\bfseries\Large XXX
            \end{minipage}%
        }%
        \endgroup
    }%
}

\begin{figure*}[t]
    \centering
    \captionsetup{skip=5pt}
    \controlledheatmap{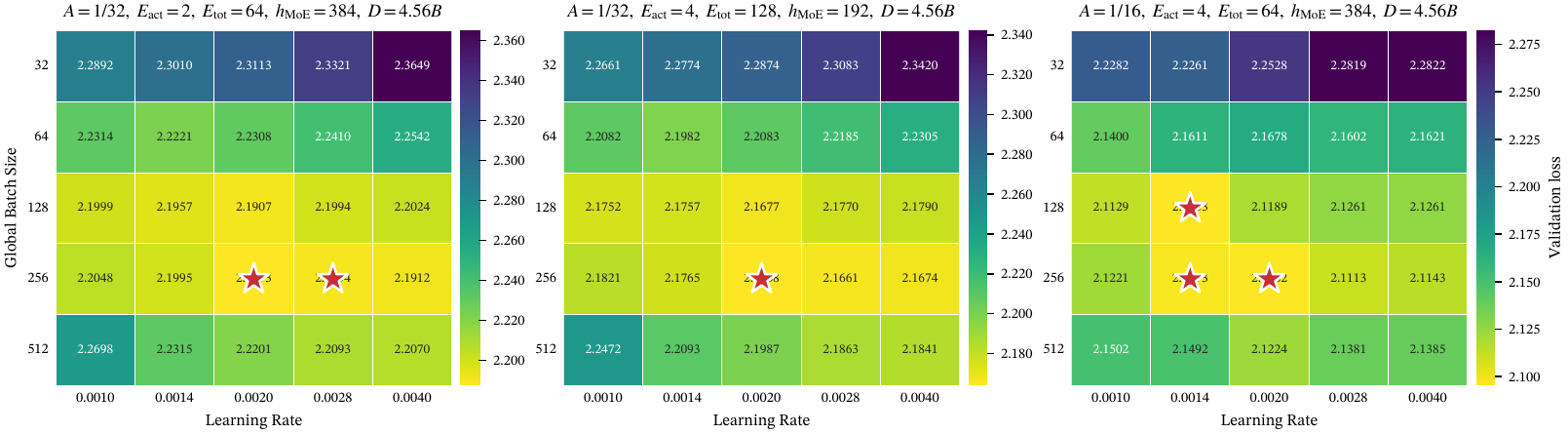}
    \caption{Expert-granularity and sparsity controls using the  10M-activated-parameter backbone configuration. Left: the reference with $A=1/32$ and $(E_{\mathrm{act}},E_{\mathrm{tot}},h_{\mathrm{MoE}})=(2,64,384)$. Middle: $(4,128,192)$ with the same $A$, activated capacity, and total capacity, testing transfer across expert granularities. Right: $(4,64,384)$ with the same total expert count $E_{tot}$, expert width, and total capacity as the reference, but $A=1/16$ due to the larger top-$n$. This complements the main sweep, which changes $A$ through $E_{tot}$. Each cell shows validation loss at one LR--BS grid point; stars mark near-optimal points.}
    \label{fig:expert-granularity-controls}
\end{figure*}

\section{Related Work}

We review two lines of related work: optimization hyperparameter scaling and sparse MoE scaling.

\paragraph*{Scaling Laws for Optimal Hyperparameters} 
Scaling laws for optimal hyperparameters aim to transfer configurations identified in small-scale experiments reliably to larger training scales. Existing studies are either theory-driven or empirically driven. Theory-driven methods, represented by the $\mu$P framework~\citep{yang2021tuning} and its extensions~\citep{yang2024tensor,dey2026don,peng2026complete}, use scale-aware parameterizations to preserve hyperparameter transfer across changes in model width, depth, and other architectural dimensions. Such approaches generally rely on particular initialization or parameterization choices and focus on variation induced by architectural scaling. Empirically driven methods instead identify and fit scale dependence directly from training experiments. They show that optimal learning rates and batch sizes vary jointly with model size and training horizon~\citep{bjorck2025scaling,bi2024deepseek,zhang2025does}; subsequent work further characterizes the coupled scaling of batch size and weight decay~\citep{bergsma2026power}. Together, these results establish predictable scaling behavior in optimization hyperparameters. We extend this empirical framework to hyperparameter scaling across MoE sparsity levels.

\paragraph*{Scaling Hyperparameters for MoEs}
Recent studies have begun to examine whether hyperparameter rules transfer from dense models to Mixture-of-Experts architectures~\citep{li2025predictable,tian2026towards,ludziejewski2025joint,zhou2026set}. Step Law reported that learning-rate and batch-size scaling laws are broadly robust across dense and MoE models~\citep{li2025predictable}. Other studies observed that MoEs tend to favor larger global batch sizes and slightly lower learning rates than comparably scaled dense models~\citep{tian2026towards,ludziejewski2025joint}. Meanwhile, work on MoE hyperparameter transfer showed that model width, depth, expert count, and expert width can be incorporated into a unified transfer parameterization~\citep{jiang2026hyperparameter}. However, existing studies largely compare dense models with a limited set of MoE configurations rather than explicitly modeling the continuous dependence of optimal hyperparameters on sparsity~\citep{tian2026towards,li2025predictable}. Some joint scaling analyses also omit relevant factors such as the training horizon~\citep{ludziejewski2025joint}, while parameterization-based transfer has been validated over only a restricted range of sparsity configurations~\citep{jiang2026hyperparameter}. Consequently, a hyperparameter scaling law that transfers systematically across sparsity levels remains missing. This gap is critical for frontier ultra-sparse MoEs~\citep{team2026kimi,xu2026deepseek}, which span much wider ranges of expert count and activation ratio. Our work systematically quantifies the effect of sparsity on optimal hyperparameters and derives unified scaling laws that transfer across sparsity levels.

\section{Limitations and Future Directions}
\label{sec:limitations}

\paragraph*{Experimental and evaluation scope.}
Our evidence comes from a single hybrid linear-attention/MLA backbone, data mixture, Muon optimizer, and sigmoid auxiliary-loss-free routing setup, while the expert-granularity and sparsity controls cover only three configurations. Transfer to other model scales, architectures, optimizers, training schedules, routing mechanisms, and broader expert configurations therefore requires further validation. We also define optimal hyperparameters using validation loss, whose improvements may not translate uniformly to downstream capabilities~\citep{gadre2025language,isik2025scaling,sun2026supervalid}. Capability-oriented objectives and data- and efficiency-scaling laws for ultra-sparse MoEs remain important directions for future work.

\paragraph*{Functional-form and extrapolation uncertainty.}
The grouped-cross-validation intervals overlap, and the number of available groups is limited; the current evidence therefore neither establishes a significant advantage for the no-interaction model nor rules out variation of the scaling exponent with activation ratio. The joint-extrapolation evaluation also contains only one target beyond the fitting ranges in $A$, $D$, and $C$, so it cannot isolate extrapolation along each variable. Although this study includes 1,800 pre-training runs, the cost of large-scale training prevents us from repeating the complete grid across multiple random seeds, as is common in lower-cost conventional machine-learning experiments. This resource constraint is broadly shared by large-scale scaling-law studies~\citep{kaplan2020scaling,hoffmann2022training,ludziejewski2025joint,tian2026towards}. Consequently, each exact prediction is evaluated once against the minimum of noisy grid observations, making the observed loss gaps descriptive point estimates. These results do not guarantee applicability to arbitrary out-of-range targets or quantify predictive uncertainty beyond the observed domain.

\section{Conclusion}

We systematically study how the optimal learning rate and batch size of ultra-sparse MoEs vary with compute scale, training horizon, and sparsity. Our central finding is that conventional scale variables alone cannot describe hyperparameter shifts across sparsity levels: the activation ratio $A$ must be treated as an additional predictive dimension. At fixed sparsity, the optimal learning rate follows a power law in training compute $C=MD$ and remains stable across different $M/D$ allocations at fixed $C$, while the optimal batch size follows a power law in training tokens $D$. Across sparsity levels, $A$ modifies the prefactors of both relations through a multiplicative power law, yielding the unified form $h^*(X,A)=k_hX^{\gamma_h}A^{\delta_h}$. Grouped prediction shows that this form outperforms the scale-only and additive alternatives. On the frozen target with 12B total parameters, the law jointly extrapolates beyond the fitting ranges in $A$, $D$, and $C$, with its prediction lying on the observed near-optimal loss plateau. Further controls show that the relation transfers across capacity-matched expert granularities. Overall, these results turn sparsity from an architectural attribute into an explicit predictor for hyperparameter selection, providing a transferable prescription for training ultra-sparse MoEs at scale.

\bibliographystyle{plainnat}
\bibliography{references}
\clearpage

\appendix

\section{Additional Experimental Details}
\label{sec:additional-experimental-details}

This section supplements the experimental setup in Section~\ref{sec:controlled-experimental-setup}. We describe the fixed training controls, followed by the model scales, compute budgets, and hyperparameter search spaces.

\subsection{Training Controls}
\label{sec:appendix-training-controls}

{Table~\ref{tab:fixed-training-controls} summarizes the training framework, tokenizer, backbone, optimization, and routing settings. All experiments use Megatron\footnote{\url{https://github.com/NVIDIA/Megatron-LM}}, 4,096-token sequences, and the same internal mixture of web, book, and code data. Engineering settings such as numerical precision and parallelism strategy are held fixed across experiments and are not treated as study variables, so they are not listed separately.}

\begin{table}[H]
    \centering
    \caption{Training controls shared across experiments.}
    \label{tab:fixed-training-controls}
    \small
    \begin{tabular}{ll}
        \toprule
        Setting & Value \\
        \midrule
        Training framework & Megatron \\
        Sequence length & 4,096 tokens \\
        Tokenizer & Byte-level BPE~\citep{shibata1999byte}, vocabulary size 157,184 \\
        Normalization & RMSNorm with QK layer normalization\\
        FFN activation & SwiGLU~\citep{shazeer2020glu} \\
        Backbone & Hybrid linear attention and multi-latent attention (MLA)~\citep{li2026ling} \\
        Position encoding & RoPE~\citep{su2024roformer} \\
        Initialization & Standard deviation 0.006; chunk initialization with $\alpha=3.0$ \\
        Optimizer & Muon~\citep{liu2025muon} \\
        LR schedule & WSD: 1\% warmup and final 10\% exponential decay~\citep{hu2024minicpm} \\
        Weight decay & 0.1, including normalization parameters \\
        Gradient clipping & Global norm 1.0 \\
        MoE routing & Sigmoid routing with Auxiliary-Loss-Free load balancing strategy~\citep{deepseekai2024deepseekv3technicalreport} \\
        \bottomrule
    \end{tabular}
\end{table}

\subsection{Model Configurations and Search Grids}
\label{sec:model-configurations}

Table~\ref{tab:model-scale-grid} summarizes the model configurations and search grid used for formal functional-family selection and coefficient fitting, together with the final held-out target with 324M activated and 12B total parameters. The formal analysis, family selection, cross-validation, and coefficient fitting use only $A\in\{1,1/4,1/16,1/32\}$ at the six activated-parameter scales, with main-grid compute budgets from $3\times10^{17}$ to $10^{20}$ FLOPs. The main grid searches $\eta\in\{5,7,10,14,20,28,40,56\}\times10^{-4}$ and $B\in\{2^{17},2^{18},\ldots,2^{23}\}$; the held-out configuration instead searches $\eta\in\{3.6,5,7,10,14,20\}\times10^{-4}$ and $B\in\{2^{19},2^{20},\ldots,2^{23}\}$. Every compute budget and Target FLOPs entry is obtained as $C=MD$ using the analytical non-embedding FLOPs/token of the corresponding configuration. All fitting scales and activation ratios share the main grid. The final target retains the observed boundary scale $N=324\mathrm{M}$, but its $A=1/64$, $D=159\mathrm{B}$, and $C=3\times10^{20}$ all lie beyond the fitting ranges, making it a single joint-extrapolation target in these three variables; this held-out configuration uses the separately specified search grid. Its losses remain uninspected until the prediction coordinates are frozen, and it enters none of the preceding operations. Under fixed top-2 routing, the five activation ratios correspond to 2, 8, 32, 64, and 128 total experts, respectively. To improve resource efficiency, we stop selected grid runs when intermediate results show that they are clearly outside the near-optimal region and unlikely to provide additional information. Accordingly, the loss surfaces in this paper denote the observed two-dimensional grids for each configuration and do not necessarily cover the full Cartesian product of the pre-specified LR--BS grid.
Compute is calculated analytically for each hybrid linear-attention/MLA MoE configuration. Specifically, $M$ counts non-embedding forward-and-backward FLOPs per token for that architecture and $C=MD$ gives non-embedding training FLOPs. The reported fits, grouped validation, fixed-$C$ slices, held-out compute, and Target FLOPs entries all use this same accounting.

\begin{table}[H]
    \centering
    \caption{Architecture controls for the six model scales. ``S/L'' gives MLA/linear-attention layers; the main grid activates two routed experts per token. Activated parameter counts and fitting roles are specified in Table~\ref{tab:model-scale-grid}.}
    \label{tab:base-model-architecture-controls}
    {\small
    \begin{tabular}{lccccc}
        \toprule
        Scale & Layers & Hidden size & Heads & \makecell{Routed-expert width ($h_{\mathrm{MoE}}$}) & S/L \\
        \midrule
        10M  & 10 & 256  & 4  & 384    & 2/8  \\
        20M  & 10 & 384  & 6  & 512    & 2/8  \\
        40M  & 12 & 512  & 8  & 640    & 3/9  \\
        80M  & 16 & 640  & 10 & 768    & 4/12 \\
        158M & 20 & 768  & 12 & 1,024  & 4/16 \\
        324M & 24 & 1,024 & 16 & 1,280 & 4/20 \\
        \bottomrule
    \end{tabular}
    }
\end{table}

\begin{table}[H]
    \centering
    \caption{Coefficient-fitting configurations and final joint-extrapolation holdout. These configurations support formal analysis, functional-family selection, grouped cross-validation, and coefficient estimation. The holdout uses a separate grid and retains the observed boundary scale $N=324\mathrm{M}$, but lies beyond the fitting ranges in $A$, $D$, and $C$; it is excluded from all four operations and evaluated only after its prediction coordinates are frozen.}
    \label{tab:model-scale-grid}
    {\small
    {%
    \begin{tabular}{lcccccccc}
        \toprule
        Scale & $N$ & $A$ & $E_{\mathrm{act}}$ & $E_{\mathrm{tot}}$ & $N_{\mathrm{tot}}$ & $D$ & \makecell{Target\\FLOPs} & \makecell{Role in\\law fitting} \\
        \midrule
        \multirow[c]{4}{*}{10M}
        & \multirow[c]{4}{*}{10M} & $1$    & $2$ & $2$  & 10M & \multirow[c]{4}{*}{4.6B} & \multirow[c]{4}{*}{$3\!\times\!10^{17}$} & \multirow[c]{4}{*}{\textbf{Fit}} \\
        & & $1/4$  & $2$ & $8$  & 27M & & & \\
        & & $1/16$ & $2$ & $32$ & 98M & & & \\
        & & $1/32$ & $2$ & $64$ & 193M & & & \\
        \midrule
        \multirow[c]{4}{*}{20M}
        & \multirow[c]{4}{*}{20M} & $1$    & $2$ & $2$  & 20M & \multirow[c]{4}{*}{7.8B} & \multirow[c]{4}{*}{$1\!\times\!10^{18}$} & \multirow[c]{4}{*}{\textbf{Fit}} \\
        & & $1/4$  & $2$ & $8$  & 55M & & & \\
        & & $1/16$ & $2$ & $32$ & 197M & & & \\
        & & $1/32$ & $2$ & $64$ & 386M & & & \\
        \midrule
        \multirow[c]{4}{*}{40M}
        & \multirow[c]{4}{*}{40M} & $1$    & $2$ & $2$  & 40M & \multirow[c]{4}{*}{11.6B} & \multirow[c]{4}{*}{$3\!\times\!10^{18}$} & \multirow[c]{4}{*}{\textbf{Fit}} \\
        & & $1/4$  & $2$ & $8$  & 111M & & & \\
        & & $1/16$ & $2$ & $32$ & 395M & & & \\
        & & $1/32$ & $2$ & $64$ & 772M & & & \\
        \midrule
        \multirow[c]{4}{*}{80M}
        & \multirow[c]{4}{*}{80M} & $1$    & $2$ & $2$  & 80M & \multirow[c]{4}{*}{19.6B} & \multirow[c]{4}{*}{$1\!\times\!10^{19}$} & \multirow[c]{4}{*}{\textbf{Fit}} \\
        & & $1/4$  & $2$ & $8$  & 223M & & & \\
        & & $1/16$ & $2$ & $32$ & 790M & & & \\
        & & $1/32$ & $2$ & $64$ & 1.5B & & & \\
        \midrule
        \multirow[c]{4}{*}{158M}
        & \multirow[c]{4}{*}{158M} & $1$    & $2$ & $2$  & 158M & \multirow[c]{4}{*}{31.7B} & \multirow[c]{4}{*}{$3\!\times\!10^{19}$} & \multirow[c]{4}{*}{\textbf{Fit}} \\
        & & $1/4$  & $2$ & $8$  & 442M & & & \\
        & & $1/16$ & $2$ & $32$ & 1.6B & & & \\
        & & $1/32$ & $2$ & $64$ & 3.1B & & & \\
        \midrule
        \multirow[c]{4}{*}{324M}
        & \multirow[c]{4}{*}{324M} & $1$    & $2$ & $2$  & 324M & \multirow[c]{4}{*}{52.9B} & \multirow[c]{4}{*}{$1\!\times\!10^{20}$} & \multirow[c]{4}{*}{\textbf{Fit}} \\
        & & $1/4$  & $2$ & $8$  & 894M & & & \\
        & & $1/16$ & $2$ & $32$ & 3.2B & & & \\
        & & $1/32$ & $2$ & $64$ & 6.2B & & & \\
        \midrule
        324M & 324M & $1/64$ & $2$ & $128$ & 12.2B & 159.0B & $3\!\times\!10^{20}$ & \textbf{Holdout} \\
        \bottomrule
    \end{tabular}
    }
    }
\end{table}

\subsection{Expert-Granularity-Controlled Configurations}
\label{sec:appendix-granularity-controlled-configuration}

The validation uses three configurations that share the approximately 10M-activated-parameter backbone, each trained on $D=4.56\mathrm{B}$ tokens. Here, ``10M'' labels the backbone configuration rather than an exactly matched activated parameter count across all three configurations. The reference $(E_{\mathrm{act}},E_{\mathrm{tot}},h_{\mathrm{MoE}})=(2,64,384)$ and the expert-granularity control $(4,128,192)$ share $A=1/32$ and match activated and total capacity. The sparsity control $(4,64,384)$ matches the reference in total expert count, expert width, $N_{\mathrm{tot}}$, and total routed capacity, but uses $A=1/16$. All configurations are excluded from coefficient estimation and functional-family selection.

\begin{table}[H]
    \centering
    \caption{Held-out expert-granularity and sparsity controls using the approximately 10M-activated-parameter backbone configuration. No row contributes to coefficient estimation or model selection.}
    \label{tab:granularity-controlled-configuration}
    \small
    \resizebox{\linewidth}{!}{%
    \begin{tabular}{ccccccc}
        \toprule
        Configuration & $(E_{\mathrm{act}},E_{\mathrm{tot}})$ & $h_{\mathrm{MoE}}$ & $A$ & Activated routed capacity & Total routed capacity & Role \\
        \midrule
        Reference & $(2,64)$ & 384 & $1/32$ & $2\times384$ & $64\times384$ & \textbf{Holdout} \\
        Granularity control & $(4,128)$ & 192 & $1/32$ & $4\times192$ & $128\times192$ & \textbf{Holdout} \\
        Sparsity control & $(4,64)$ & 384 & $1/16$ & $4\times384$ & $64\times384$ & \textbf{Holdout} \\
        \bottomrule
    \end{tabular}
    }
\end{table}

\section{Additional Experimental Results}
\label{sec:additional-experimental-results}

{This section reports supplementary sparsity slices and fitted coefficients for the candidate functional families.}

\subsection{Sparsity Dependence Across Fixed-Scale Slices}
\label{sec:appendix-sparsity-slices}

The right panels of Figures~\ref{fig:sparsity-dependencies-lr} and~\ref{fig:sparsity-dependencies-bs} show the dependence of the near-optimal hyperparameters on the activation ratio at one fixed compute or token budget, respectively. Here, we extend this analysis to three fixed-$C$ slices and three fixed-$D$ slices to test whether the relationship persists across training scales.
Across all slices, the sparsity-induced shifts are not specific to a single reference scale: over the compute and token ranges examined, both near-optimal learning rates and batch sizes follow the separable multiplicative power-law correction $A^{\delta_h}$ used in the main text.

\begin{figure*}[t]
    \centering
    \includegraphics[width=\textwidth]{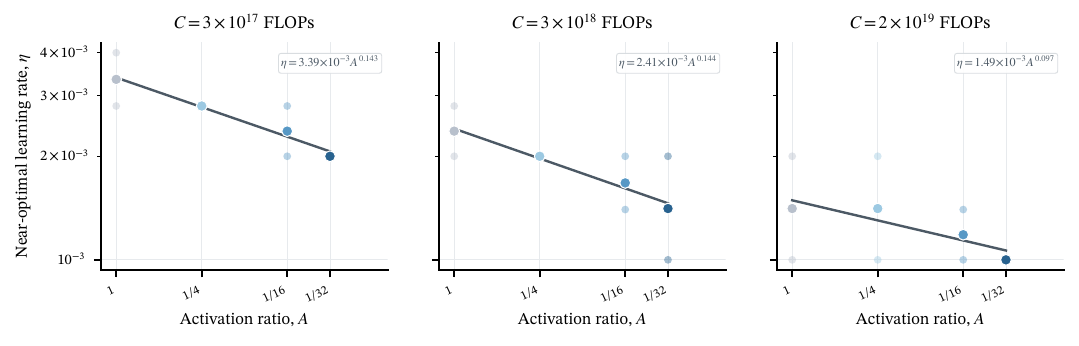}
    \caption{Learning-rate dependence on sparsity across fixed-compute slices. Each panel shows near-optimal learning rates as a function of $A$ at a fixed compute budget: $C=3\times10^{17}$, $3\times10^{18}$, and $2\times10^{19}$ FLOPs from left to right. Faint markers denote individual observations, highlighted markers their geometric means, and solid lines show multiplicative power-law fits in $A$.}
    \label{fig:appendix-lr-sparsity-slices}
\end{figure*}

\begin{figure*}[t]
    \centering
    \includegraphics[width=\textwidth]{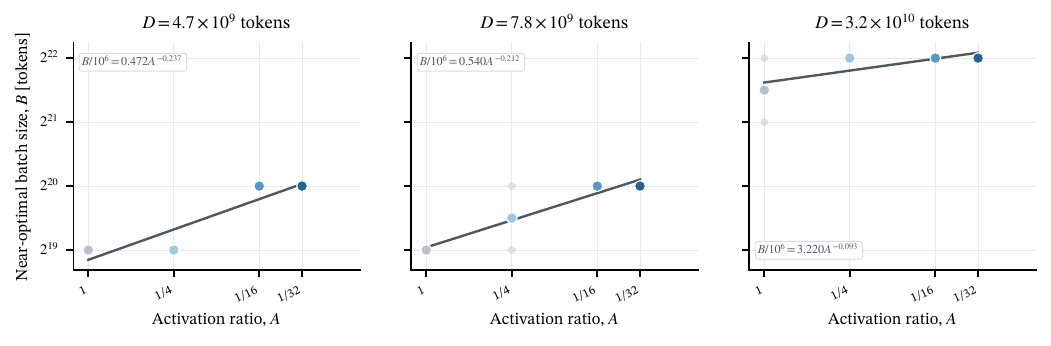}
    \caption{Batch-size dependence on sparsity across fixed-token slices. Each panel shows near-optimal batch sizes as a function of $A$ at a fixed training horizon: $D=4.7\times10^9$, $7.8\times10^9$, and $3.2\times10^{10}$ tokens from left to right. Faint markers denote individual observations, highlighted markers their geometric means, and solid lines show multiplicative power-law fits in $A$.}
    \label{fig:appendix-bs-sparsity-slices}
\end{figure*}

\subsection{{Candidate Functional-Form Coefficients}}
\label{sec:appendix-extrapolation-ablation}

{Table~\ref{tab:appendix-extrapolation-ablation} reports the full-development-set coefficients and log-space fit metrics for all four candidate families. RMSE and $R^2$ are computed from base-2 log residuals on the same target-aligned surfaces, with lower RMSE and higher $R^2$ indicating better in-sample fit. These descriptive metrics are distinct from the grouped out-of-fold errors used for model selection in the main text. Although the additive family achieves the best in-sample metrics, the multiplicative law is selected because its grouped predictive performance is comparable while using one fewer parameter and admitting a simpler interpretation.}
\begin{table}[H]
    \centering
    \caption{Fitted coefficients and in-sample fit metrics for the four candidate families. RMSE and $R^2$ are evaluated in base-2 log space on the target-aligned development surfaces. Lower RMSE and higher $R^2$ indicate better in-sample fit.}
    \label{tab:appendix-extrapolation-ablation}
    \resizebox{\textwidth}{!}{%
    \begin{tabular}{llllrr}
        \toprule
        Candidate family & Functional form & Target & Fitted coefficients & RMSE $\downarrow$ & $R^2$ $\uparrow$ \\
        \midrule
        \multirow{2}{*}{Scale only}
        & \multirow{2}{*}{$kX^\gamma$} & LR ($X=C$)
        & $k=0.8704,\ \gamma=-0.1463$ & 0.3634 & 0.3879 \\
        & & BS ($X=D$)
        & $k=1.4827,\ \gamma=0.5894$ & 0.3057 & 0.7104 \\
        \addlinespace
        \multirow{2}{*}{Additive}
        & \multirow{2}{*}{$k_X X^\gamma+k_A A^\delta$} & LR ($X=C$)
        & $k_X=1.7579\!\times\!10^8,\ \gamma=-0.6362,\ k_A=1.9358\!\times\!10^{-3},\ \delta=0.1808$ & \textbf{0.2358} & \textbf{0.7422} \\
        & & BS ($X=D$)
        & $k_X=0.0197,\ \gamma=0.7651,\ k_A=1.1802\!\times\!10^5,\ \delta=-0.3412$ & \textbf{0.2644} & \textbf{0.7834} \\
        \addlinespace
        \multirow{2}{*}{Log interaction}
        & \multirow{2}{*}{$kX^{\beta_X+\beta_{XA}\log_2 A}A^{\beta_A}$} & LR ($X=C$)
        & $k=0.3173,\ \beta_X=-0.1155,\ \beta_A=-0.1434,\ \beta_{XA}=0.0046$ & 0.2452 & 0.7214 \\
        & & BS ($X=D$)
        & $k=0.1271,\ \beta_X=0.6896,\ \beta_A=-1.1308,\ \beta_{XA}=0.0319$ & 0.2661 & 0.7805 \\
        \addlinespace
        \midrule
        \multirow{2}{*}{Multiplicative (selected)}
        & \multirow{2}{*}{$kX^\gamma A^\delta$} & LR ($X=C$)
        & $k=0.8343,\ \gamma=-0.1385,\ \delta=0.1361$ & 0.2463 & 0.7188 \\
        & & BS ($X=D$)
        & $k=6.4765,\ \gamma=0.5181,\ \delta=-0.0841$ & 0.2765 & 0.7630 \\
        \bottomrule
    \end{tabular}%
    }
\end{table}

\section{Theoretical Analysis}
\label{sec:appendix-theoretical-analysis}

This section gives a simple local model for the two sparsity trends observed in the main text: as the activation ratio $A$ decreases, the optimal batch size increases and the optimal learning rate decreases. We use the notation of the main text, where $E_{\mathrm{act}}$ and $E_{\mathrm{tot}}$ denote the number of experts activated per token and the total number of experts, respectively, and $A=E_{\mathrm{act}}/E_{\mathrm{tot}}$.

\subsection{Why Sparsity Increases the Optimal Batch Size}
\label{app:bs-sparsity-theory}

Consider an MoE layer with global token batch size $B$. Under balanced routing, each expert receives approximately $AB$ tokens on average. Under an independent-sample approximation, the variance of its gradient estimate is inversely proportional to this effective batch size:
\begin{equation}
    \operatorname{Var}(\widehat{g}_e)
    \propto \frac{1}{AB}.
    \label{eq:expert-gradient-variance}
\end{equation}
Thus, decreasing $A$ increases expert-side gradient noise and favors a larger global batch. At a fixed training-token budget $D$, however, increasing $B$ reduces the number of optimizer updates to $D/B$. Meanwhile, shared parameters still receive gradients from the full batch and are not thinned by sparse routing. We model this trade-off with the local approximation
\begin{equation}
    \Delta\mathcal{L}(B;D,A)
    \simeq
    c_{\mathrm{step}}\left(\frac{B}{D}\right)^p
    +c_{\mathrm{shared}}B^{-q}
    +c_{\mathrm{expert}}(AB)^{-q},
    \qquad p,q>0,
    \label{eq:batch-size-local-loss}
\end{equation}
where all three coefficients are positive. The terms represent the penalties from fewer optimizer updates, gradient noise in shared parameters, and gradient noise in expert parameters, respectively. Minimizing over $B$ gives
\begin{equation}
    B^*(D,A)
    =
    \left[
        \frac{q\left(c_{\mathrm{shared}}
        +c_{\mathrm{expert}}A^{-q}\right)}
        {p\,c_{\mathrm{step}}}
    \right]^{\frac{1}{p+q}}
    D^{\frac{p}{p+q}}.
    \label{eq:optimal-bs-sparsity}
\end{equation}
In this model, the training-token exponent is $\gamma_B=p/(p+q)$ and does not depend on $A$, consistent with the shared-exponent form used in the main text. The local elasticity with respect to $A$ is
\begin{equation}
    \delta_B(A)
    \equiv
    \frac{\partial\log B^*}{\partial\log A}
    =
    -\frac{q}{p+q}
    \frac{c_{\mathrm{expert}}A^{-q}}
    {c_{\mathrm{shared}}+c_{\mathrm{expert}}A^{-q}},
    \qquad -1<\delta_B(A)<0.
    \label{eq:local-bs-sparsity-exponent}
\end{equation}
Thus, the optimal batch size increases as $A$ decreases, but more slowly than $A^{-1}$. The update-count penalty and shared-parameter noise both weaken the effect of sparse routing, consistent with the sublinear trend in Figure~\ref{fig:appendix-bs-sparsity-slices}.

\subsection{Why Sparsity Decreases the Optimal Learning Rate}
\label{app:lr-sparsity-theory}

Equation~\eqref{eq:expert-gradient-variance} shows that expert-side gradient noise increases as the effective expert batch $AB$ decreases. Under a local noise-scale approximation, the expert-side noise induced by one optimizer update scales as
\begin{equation}
    \mathcal{T}_e \propto \frac{\eta}{AB},
    \label{eq:expert-update-noise}
\end{equation}
where $\eta$ is the learning rate. A smaller effective expert batch therefore favors a smaller optimal learning rate. We summarize the response of the optimizer and the buffering effect of shared parameters with an exponent $\rho$:
\begin{equation}
    \eta^*\propto (AB^*)^\rho,
    \qquad 0<\rho\leq 1.
    \label{eq:lr-effective-expert-batch}
\end{equation}
From Equation~\eqref{eq:local-bs-sparsity-exponent}, the increase in $B^*$ is not large enough to offset the decrease in $A$, because
\begin{equation}
    \frac{\partial\log(AB^*)}{\partial\log A}
    =1+\delta_B(A)>0.
\end{equation}
The corresponding local sparsity exponent for the learning rate is therefore
\begin{equation}
    \delta_\eta(A)
    \equiv
    \frac{\partial\log\eta^*}{\partial\log A}
    =\rho\bigl(1+\delta_B(A)\bigr)>0.
    \label{eq:local-lr-sparsity-exponent}
\end{equation}
As $A$ decreases, the effective number of samples received by each expert still falls, so the optimal learning rate also decreases. The accompanying increase in $B^*$ partially offsets this effect, while shared parameters are not directly thinned by sparse routing; the resulting learning-rate shift is therefore typically modest.

The analysis relies on simplifying assumptions, including balanced routing, independent samples, and a local noise model. It explains the empirical signs $\delta_B<0$ and $\delta_\eta>0$ and the approximately shared base exponent for batch size across sparsity levels; it does not derive the fitted coefficients from first principles or require the sparsity exponents to be exactly constant. In particular, the base scaling of the learning rate with compute $C$ remains determined empirically in the main text.

\end{document}